\documentclass[3p,twocolumn]{elsarticle}

\usepackage{amssymb}
\usepackage{graphicx}
\usepackage{subcaption}
\usepackage{balance}
\usepackage{amsthm}

\journal{Computer, Speech \& Language}

\begin{document}

\begin{frontmatter}



\title{Articulatory and Bottleneck Features for \\ Speaker-Independent ASR of Dysarthric Speech}
\tnotetext[mytitlenote]{The second author is currently working with Apple Inc. This research is funded by the NWO Project 314-99-101 (CHASING) and
NWO Project 314-99-119 (Frisian Audio Mining Enterprise).}

\author[nus,run]{Emre Y\i lmaz\corref{cor1}}
\ead{emre@nus.edu.sg}
\author[uom]{Vikramjit Mitra}
\ead{vmitra@umd.edu}
\author[pin]{Ganesh Sivaraman}
\ead{ganesa90@gmail.com}
\author[sri]{Horacio Franco}
\ead{horacio.franco@sri.com}
\cortext[cor1]{Corresponding author, tel. +65-9242-5322}
\address[nus]{Dept. of Electrical and Computer Engineering, National University of Singapore, Singapore}
\address[run]{CLS/CLST, Radboud University, Nijmegen, Netherlands}
\address[uom]{University of Maryland, College Park, MD, USA}
\address[pin]{Pindrop, Atlanta, USA}
\address[sri]{STAR Laboratory, SRI International, Menlo Park, CA, USA}


\author{}

\address{}

\begin{abstract}
The rapid population aging has stimulated the development of assistive devices that provide personalized medical support to the needies suffering from various etiologies. One prominent clinical application is a computer-assisted speech training system which enables personalized speech therapy to patients impaired by communicative disorders in the patient's home environment. Such a system relies on the robust automatic speech recognition (ASR) technology to be able to provide accurate articulation feedback. With the long-term aim of developing off-the-shelf ASR systems that can be incorporated in clinical context without prior speaker information, we compare the ASR performance of speaker-independent bottleneck and articulatory features on dysarthric speech used in conjunction with dedicated neural network-based acoustic models that have been shown to be robust against spectrotemporal deviations. We report ASR performance of these systems on two dysarthric speech datasets of different characteristics to quantify the achieved performance gains. Despite the remaining performance gap between the dysarthric and normal speech, significant improvements have been reported on both datasets using speaker-independent ASR architectures.
\end{abstract}


\end{frontmatter}


\section{Introduction}
\label{sec:intro}

Among the problems that are likely to be associated with an increasingly ageing population worldwide is a growing incidence of neurological disorders such as Parkinson's disease (PD), cerebral vascular accident (CVA or stroke) and traumatic brain injury (TBI). Possible consequences
of such diseases are motor speech disorders including dysarthria caused by neuromuscular control problems \cite{duffy1995} which often lead to decreased speech intelligibility and communication impairment \cite{kent2003}. As a result, the life quality of dysarthric patients is negatively affected \cite{walshe2011} and they run the risk of losing contact with friends and relatives and eventually becoming isolated from the society.

Research has shown that intensive therapy can be effective in (speech) motor rehabilitation \cite{ramig2001,bhogal2003,kwakkel2006,rijntjes2009}, but various factors conspire to make intensive therapy expensive and difficult to obtain. Recent developments show that therapy can be provided without resorting to frequent face-to-face sessions with therapists by employing computer-assisted speech training systems \cite{beijer2011}. According to the outcomes of the efficacy tests presented in \cite{beijer2014}, the user satisfaction appears to be quite high. However, most of these systems are not yet capable of automatically assessing the articulation accuracy at the level of individual speech sounds, which are known to have an impact on speech intelligibility \cite{debodt2002,yunusova2005,nuffelen2009,popovici2012,ganzeboom2016}.

In this work, we describe our efforts to develop an ASR framework that is robust to variations in pathological speech without relying on speaker information and can be incorporated in various clinical applications including personalized speech therapy tools. In particular, multiple speaker-independent ASR architectures are presented that are designated to perform robust ASR in the presence of spectrotemporal deviations in speech. Training robust acoustic models to capture the within- and between-speaker variation in dysarthric speech is generally not feasible due to the limited size and structure of existing pathological speech databases. The number of recordings in dysarthric speech databases is much smaller compared to that in normal speech databases. Despite long-lasting efforts to build speaker- and text-independent ASR systems for people with dysarthria, the performance of state-of-the-art systems is still considerably lower on this type of speech than on normal speech \cite{sanders2002,hasegawa2006,rudzicz2007,caballero2009,mengistu2011,seong2012,christensen2012,shahamiri2014,takashima2015,lee2016,joy2017}.

In previous work \cite{yilmaz2016_5}, we described a solution to train a better deep neural network (DNN)-hidden Markov model (HMM) system for the Dutch language, a language that has fewer speakers and resources compared to English. In particular, we investigated combining non-dysarthric speech data from different varieties of the Dutch language to train more reliable acoustic models for a DNN-HMM ASR system. This work was conducted in the framework of the CHASING project\footnote{http://hstrik.ruhosting.nl/chasing/}, in which a serious game employing ASR is being developed to provide additional speech therapy to dysarthric patients \cite{ganzeboom2016_2}. Moreover, we created a 6-hour Dutch dysarthric speech database that had been collected in a previous project (EST) \cite{yilmaz2016_2} for training purposes and investigate the impact of multi-stage DNN training (henceforth referred to as model adaptation) for pathological speech~\cite{yilmaz2017}.

One way of addressing the challenges introduced due to dysarthria is to include additional several articulatory features (AF) which contain information about the shape and dynamics of the vocal tract given that such information has been shown to be directly related with the spectro-temporal deviations in pathological speech (\cite{walsh2012} and references therein). Using AFs together with acoustic features has been investigated and shown to be beneficial in the ASR of normal speech, e.g. \cite{zlokarnik1995,wrench2000,stephenson2000,markov2006,mitra2012,badino2016}. A subset of these approaches learn a mapping between acoustic and articulatory spaces for the speech inversion, and use the learned articulatory information in an ASR system for improved representation of speech in a high-dimensional feature space. Rudzicz \cite{rudzicz2011} tried using discrete AF together with conventional acoustic features for phone classification experiments on dysarthric speech. 

More recently, \cite{mitra2017} has proposed the use of convolutional neural networks (CNN) for learning speaker independent articulatory models for mapping acoustic features to the corresponding articulatory space. Later, a novel acoustic model is designed to integrate the AF together with the acoustic features~\cite{mitra2017_2}. Motivated by the success of these systems, we have investigated the joint use of articulatory and acoustic features for the speaker-independent ASR of pathological speech and reported promising initial results~\cite{yilmaz2018}. Specifically, the use of vocal tract constriction variables (TVs) and mel filterbank features as input to fused-feature-map CNN (fCNN) acoustic models~\cite{mitra2017_2} has been compared with other deep and (frequency and time-frequency) convolutional neural network architectures~\cite{mitra2015}.

In this paper, we firstly explore the use of gammatone by comparing the ASR performance to the conventional mel filterbank features in the context of dysarthric speech recognition. The gammatone filters can capture fine-resolution changes in the spectra and are closer to the auditory filterbanks formulated through perceptual studies compared to the mel filters. In this sense, the mel filters can be considered as an approximation of the gammatone filters. Moreover, we extend our previous work \cite{yilmaz2018} by concatenating articulatory features with gammatone filterbank acoustic features using speech inversion systems trained on both synthetic and real speech. It is vital to note that the type of articulatory features used in this work is novel, as prior work focuses on discrete articulatory features, whereas the articulatory features in this work are continuous and represent vocal tract kinematics. The rationale behind using these articulatory representations is to capture the speech production space which helps to account for any variability observed in the acoustic space, hence making the feature combination robust. To the best of authors' knowledge, this is the first work to explore articulatory features, which are estimated using speech inversion models trained on synthetic and real speech, in various capacities for dysarthric speech recognition. Similar articulatory representations have been used successfully before to account for acoustic variations such as coarticulation and non-native speech~\cite{mitra2017_2} and noise and channel variations~\cite{mitra2017}.

Another effective way of representing speech is achieved by using bottleneck features extracted from a bottleneck layer which can be considered a fixed dimensional non-linear transformation which gives a stable representation of the sub-word unit subspace given acoustics. Such representations have been found to be useful in low-resourced language scenarios in which it is not viable to accurately represent the complete phoneme space due to data scarcity. The target low-resourced language benefits from additional speech data from other high-resourced languages possibly sharing similar acoustic units \cite{vu2012b}. In case of dysarthric speech, using bottleneck features has been shown to mitigate the effect of the increased variations observed as a result of poor articulation skills thanks to their disorder invariant characteristics ~\cite{nakashika2014,takashima2015}. In this work, we extend this analysis to (1) the novel time-frequency convolutional neural network (TFCNN) framework aiming to obtain improved bottleneck representation of dysarthric speech and (2) investigate the impact of applying the model adaptation technique described in Section \ref{ssec:bn-ma} at different stages aiming to discover the systems providing bottleneck features with increased robustness against the variations in pathological speech.

A list of the contributions of this submission is given below for clarity: 
\begin{enumerate}
	\item A comparison of the use of gammatone and mel filterbank features is given in the context of dysarthric speech recognition.
	\item We investigate the use of continuous articulatory features representing vocal tract kinematics by concatenating them with acoustic features where the speech inversion system is trained using synthetic and real speech.
	\item The ASR performance of the bottleneck features extracted using the designated convolutional NN models is explored by jointly learning the bottleneck layers and feature map fusion.
	\item For the scenario with training dysarthric speech data, the impact of model adaptation is explored on both acoustic models and bottleneck extraction schemes.  
\end{enumerate}

This paper is organized as follows. Section~\ref{sec:relwork} summarizes some previous literature on the ASR of dysarthric speech. Section~\ref{sec:scs} details the speech corpora used in this work by providing some information about the speakers for each dysarthric speech database. Proposed speaker-independent ASR schemes are described in Section~\ref{sec:siasr}. The experimental setup is described in Section~\ref{sec:expset}, and the recognition results are presented in Section~\ref{sec:res}. A general discussion is given in Section~\ref{sec:dis} before concluding the paper in Section~\ref{sec:conc}.

\section{Related work}
\label{sec:relwork}

\begin{table*}
\centering
\caption{Summary of the speech resources used in the ASR experiments (h:hours, m:minutes)}
\begin{tabular}{| l | c | c | c | c |} 
\hline
Speech database   & Type       & Train   & Test   \\
\hline\hline
CGN Dutch     & Normal     &   255h   &   -  \\
\hline
CGN Flemish   & Normal     &   186h 30m  &   -  \\
\hline
EST Dutch~\cite{yilmaz2016_2} & Dysarthric &   4h 47m &  -    \\
\hline
CHASING01 Dutch~\cite{yilmaz2017} & Dysarthric &     -   &  55m     \\
\hline
COPAS Flemish~\cite{middagphd}& Dysarthric &     -   &  1h 30m \\
\hline
COPAS Flemish~\cite{middagphd}& Normal (Control) &  -   &  1h \\
\hline
\end{tabular}
\label{tab:data_summary}
\end{table*}

There have been numerous efforts to build ASR systems operating on pathological speech. Lee et al. \cite{lee2016} has reported the ASR performance on Cantonese aphasic speech and disordered voice. A DNN-HMM system provided significant improvements on disordered voice and minor improvements on aphasic speech compared to a GMM-HMM system. Takashima et al. \cite{takashima2015} proposed a feature extraction scheme using convolutional bottleneck networks for dysarthric speech recognition. They tested the proposed approach on a small test set consisting of 3 repetitions of 216 words by a single male speaker with an articulation disorder and reported some gains over a system using MFCC features. In a recent work, Kim et al. \cite{kim2018} investigated convolutional long short-term memory (LSTM) networks dysarhtric speech from 9 speakers. They reported improved ASR accuracies compared to CNN- and LSTM-based acoustic models.

Shahamiri and Salim \cite{shahamiri2014} proposed an artificial neural network-based system trained on digit utterances from nine non-dysarthric and 13 dysarthric individuals affected by Cerebral Palsy (CP). Christensen et al. \cite{christensen2012} trained their models solely on 18 hours of speech of 15 dysarthric speakers due to CP leaving one speaker out as test set. Rudzicz \cite{rudzicz2007} compared the performance of a speaker-dependent and a speaker-adaptive GMM-HMM systems on the Nemours database \cite{menendez1996}. Later, Rudzicz \cite{rudzicz2011} tried using AF together with conventional acoustic features for phone classification experiments on dysarhtric speech. Mengistu and Rudzicz \cite{mengistu2011} combined dysarthric data of eight dysarthric speakers with that of seven normal speakers, leaving one out as test set and obtained an average increase by 13.0\% in comparison to models trained on non-dysarthric speech only. In another work \cite{mengistu2011_2}, they compared the recognition performance of n{\"a}ive human listener and ASR systems. 

In one of the earliest work on Dutch pathological speech by Sanders et al. \cite{sanders2002}, a pilot study was presented on ASR of Dutch dysarthric speech data obtained from two speakers with a birth defect and a cerebrovascular accident. Both speakers were classified as mild dysarthric by a speech pathologist. Seong et al. \cite{seong2012} proposed a weighted finite state transducer (WSFT)-based ASR correction technique applied to an ASR system trained. Similar work had been proposed by Caballero-Morales and Cox~\cite{caballero2009} previously.

In the scope of the CHASING Project, we have been developing a serious game employing ASR to provide additional speech therapy to dysarthric patients~\cite{ganzeboom2016, ganzeboom2016_2}. Earlier research in this project is reported in~\cite{yilmaz2016_5} and~\cite{yilmaz2017} in which we focus on investigating the available resources by using speech data from different varieties of the Dutch language and investigate model adaptation to tune an acoustic model using a small amount of dysarthric speech for training purposes respectively.

In general, it is difficult to compare results between these publications due to the differences in types of speech materials, types of dysarthria, reported severity, and datasets used for training and testing. Additionally, dysarthric speech is highly variable in nature, not only due to its various etiologies and degrees of severity, but also because of possible individually deviating speech characteristics. This may negatively influence the capability of speaker-independent systems to generalize over multiple dysarthric speakers.

\section{Speech corpora selection}
\label{sec:scs}

We use Dutch and Flemish dysarthric speech corpora for investigating the effectiveness of different ASR systems. The details of the speech data used in this work are presented in Table \ref{tab:data_summary}. There are two dysarthric speech corpora available in Dutch, namely EST~\cite{yilmaz2016_2} and CHASING01~\cite{yilmaz2017}, which are used for training and testing purposes respectively. For Flemish, we only have the COPAS corpus~\cite{middagphd} which has enough speech data for testing only. Given the limited availability of dysarthric speech data, we also use already existing databases of normal Dutch and Flemish speech to train acoustic models. In the following paragraphs, we detail the normal and dysarthric speech corpora used in this work.

There have been multiple Dutch-Flemish speech data collection efforts~\cite{cgn,jasmin} which facilitate the integration of both Dutch and Flemish data in the present research. For training purposes, we used the \textit{Corpus Gesproken Nederlands} (CGN) corpus~\cite{cgn}, which contains representative collections of contemporary standard Dutch as spoken by adults in the Netherlands and Flanders. The CGN components with read speech, spontaneous conversations, interviews and discussions are used for acoustic model training. The duration of the normal Flemish (VL) and northern Dutch (NL) speech data used for training is 186.5 and 255 hours, respectively.

The EST Dutch dysarthric speech database contains dysarthric speech from ten patients with Parkinson's Disease (PD), four patients who have had a Cerebral Vascular Accident (CVA), one patient who suffered Traumatic Brain Injury (TBI) and one patient having dysarthria due to a birth defect. Based on the meta-information, the age of the speakers is in the range of 34 to 75 years with a median of 66.5 years. The level of dysarthria varies from mild to moderate.

The dysarthric speech collection for this database was achieved in several experimental contexts. The speech tasks presented to the patients in these contexts consist of numerous word and sentence lists with varying linguistic complexity. The database includes 12 Semantically Unpredictable Sentences (SUSs) with 6- and 13-word declarative sentences, 12 6-word interrogative sentences, 13 Plomp and Mimpen sentences, 5 short texts, 30 sentences with /t/, /p/ and /k/ in initial position and unstressed syllable, 15 sentences with /a/, /e/ and /o/ in unstressed syllables, production of 3 individual vowels /a/, /e/ and /o/, 15 bisyllabic words with /t/, /p/ and /k/ in initial position and unstressed syllable and 25 words with alternating vowel-consonant composition (CVC, CVCVCC, etc.).

The EST database contains 6 hours and 16 minutes of dysarthric speech material from 16 speakers~\cite{yilmaz2016_2}. The speech segments with pronunciation error marks indicating what is said by the speaker either (1) obviously differs from the reference transcription or (2) is unintelligible, were excluded from the training set to ensure a high degree of transcription quality.  Additionally, the segments including a single word and pseudoword were also excluded, since the sentence reading tasks are more relevant in this scenario. The total duration of the dysarthric speech data eventually selected for training is 4 hours and 47 minutes.

For testing purposes, we firstly use the sentence reading tasks of the CHASING01 Dutch dysarthric speech database. This database contains speech of 5 patients who participated in speech training experiments and were tested at 6 different times during the treatment. For each set of audio files, the following material was collected: 12 SUSs, 30 /p/, /t/, /k/ sentences in which the first syllable of the last word is unstressed and starts with /p/, /t/ or /k/, 15 vowel sentences with the vowels /a/,/e/ and /o/ in stressed syllables, appeltaarttekst (\textit{apple cake recipe}) in 5 parts. Utterances that deviated from the reference text due to pronunciation errors (e.g. restarts, repeats, hesitations, etc.) were removed. After this subselection, the utterances from 3 male patients remained and were included in the test set. These speakers are 67, 62 and 59 years old, two of them having PD and the third having had a CVA. This database contains 721 utterances (in total 6231 words) spoken by 3 dysarthric speakers with a total duration of 55 minutes.

All sentence reading tasks with annotations from the Flemish COPAS pathological speech corpus, namely 2 isolated sentence reading tasks, 11 text passages with reading level difficulty of AVI 7 and 8 and Text Marloes, are also included as a second test set. The COPAS corpus contains recordings from 122 Flemish normal speakers and 197 Flemish speakers with speech disorders such as dysarthria, cleft, voice disorders, laryngectomy and glossectomy. The dysarthric speech component contains recordings from 75 Flemish patients affected by Parkinson's disease, traumatic brain injury, cerebrovascular accident and multiple sclerosis who exhibit dysarthria at different levels of severity. Containing more speakers with more diverse etiologies, performing ASR on this corpus is found to more challenging compared to the CHASING01 dysarthric speech database (c.f. the ASR results in~\cite{yilmaz2016_5,yilmaz2017,yilmaz2018}). 212 different sentence tasks from the COPAS database are included in the experiments which are uttered by 103 dysarthric and 82 normal speakers. The sentence tasks uttered in the Flemish corpus by normal speakers (SentNor) and speakers with disorders (SentDys) consists of 1918 (15,149) and 1034 (8287) sentences (words) with a total duration of 1.5 and 1 hour, respectively.

\begin{figure*}[t]
  \vspace{-1cm}
  \begin{subfigure}[t]{1.0\textwidth}
  \centering
  \includegraphics[width=5.4in]{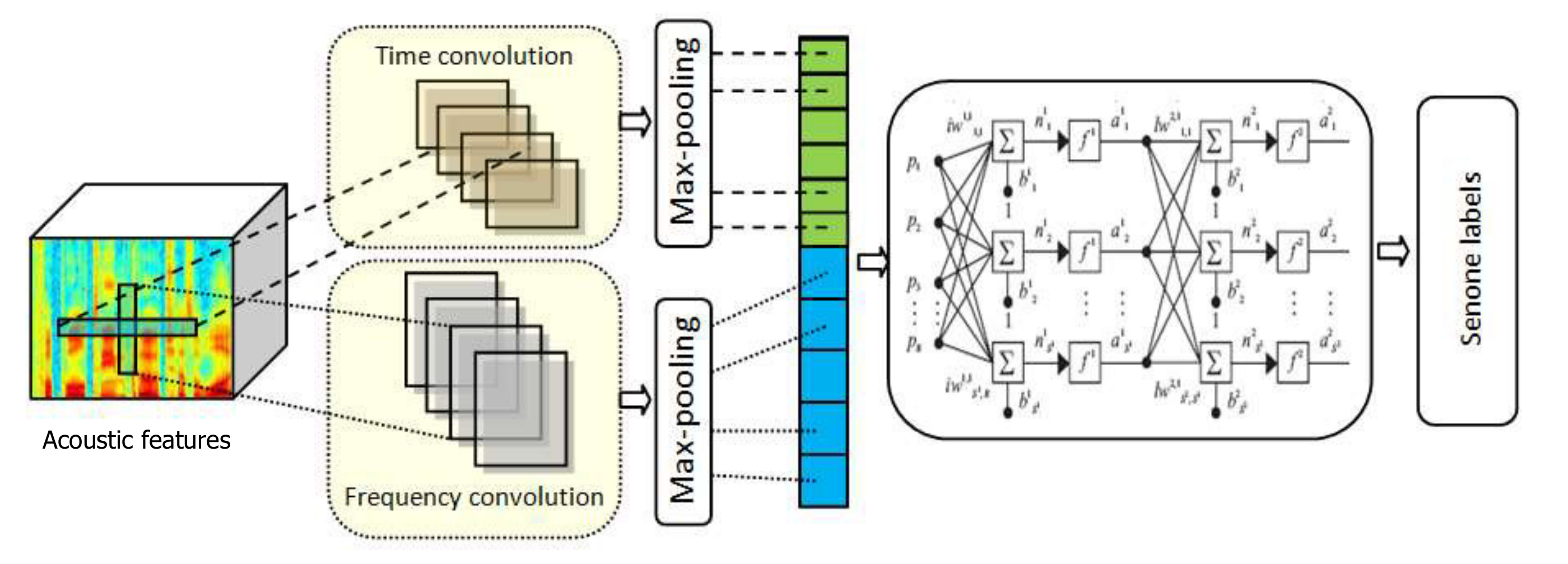}
  \caption{A time-frequency convolutional neural network (TFCNN)~\cite{mitra2015}}
  \label{fig:tfcnn}
  \end{subfigure}
  \centering
  \begin{subfigure}[t]{1.0\textwidth}
  \centering
  \includegraphics[width=5.4in]{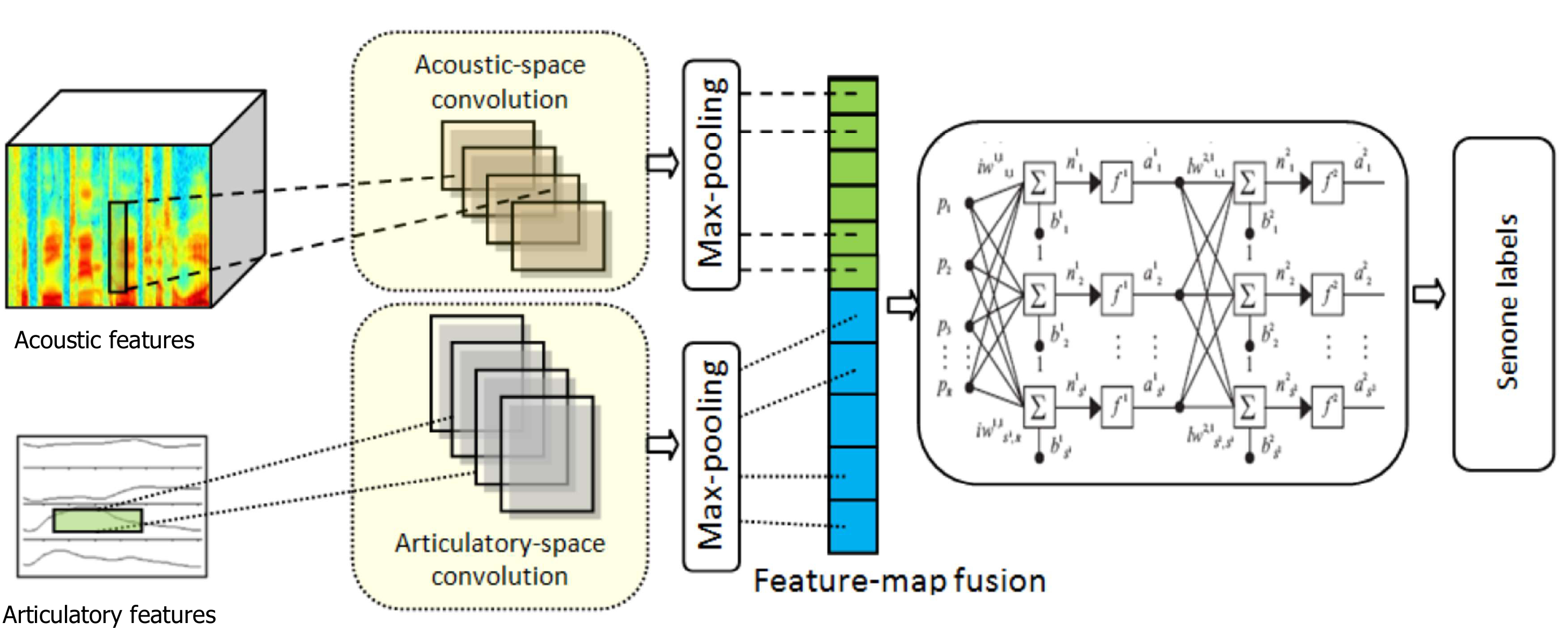}
  \caption{A fused-feature-map convolutional neural network (fCNN)~\cite{mitra2017_2}}
  \label{fig:fcnn}
  \end{subfigure}
  \vspace{-0.15cm}
  \caption{Designated convolutional neural network architectures with increased robustness against spectrotemporal deviations}
  \vspace{-0.5cm}
\end{figure*}

\section{Speaker-independent ASR Schemes for dysarthric speech}
\label{sec:siasr}

\subsection{Acoustic models}
\label{ssec:acmod}

In the scope of this work, we have trained several NN-based acoustic models to investigate their performance on dysarthric speech using different speaker-independent features. Standard DNN and CNN models trained on mel- and gammatone filterbanks are compared with novel NN architectures such as time-frequency convolutional nets (TFCNN) and a fused-feature-map convolutional neural network (fCNN) which are detailed in the following paragraphs. 

Time-frequency convolutional nets (TFCNN) are a suitable candidate for the acoustic modeling of dysarthric speech. As depicted in Figure \ref{fig:tfcnn}, TFCNN is a modified convolution network in which two parallel layers of convolution filters operate on the input feature space: one across time (time-convolution) and the other across frequency (frequency-convolution). The feature maps from these two convolution layers are fused and then fed to a fully connected deep neural network. Performing convolution both on time and frequency axes, they exhibit increased robustness against the spectrotemporal deviations due to background noise~\cite{mitra2015}.

The acoustic and articulatory (concatenated) features are fed to a fCNN which is illustrated in Figure \ref{fig:fcnn}. This architecture uses two types of convolutional layers. The first convolutional layer operates on the acoustic features, which are the filterbank energy features, and performs convolution across frequency. The other convolutional layer operates on AFs, which are the TV trajectories, and performs convolution across time. The output of the max-pooling layers are fed to a single NN after performing feature-map fusion. AFs are used only in conjunction with the fCNN models, as the other NN architectures are found to provide worse recognition performance using the concatenated features~\cite{mitra2017_2}.
\begin{figure*}
  \centering
  \includegraphics[width=6.5in]{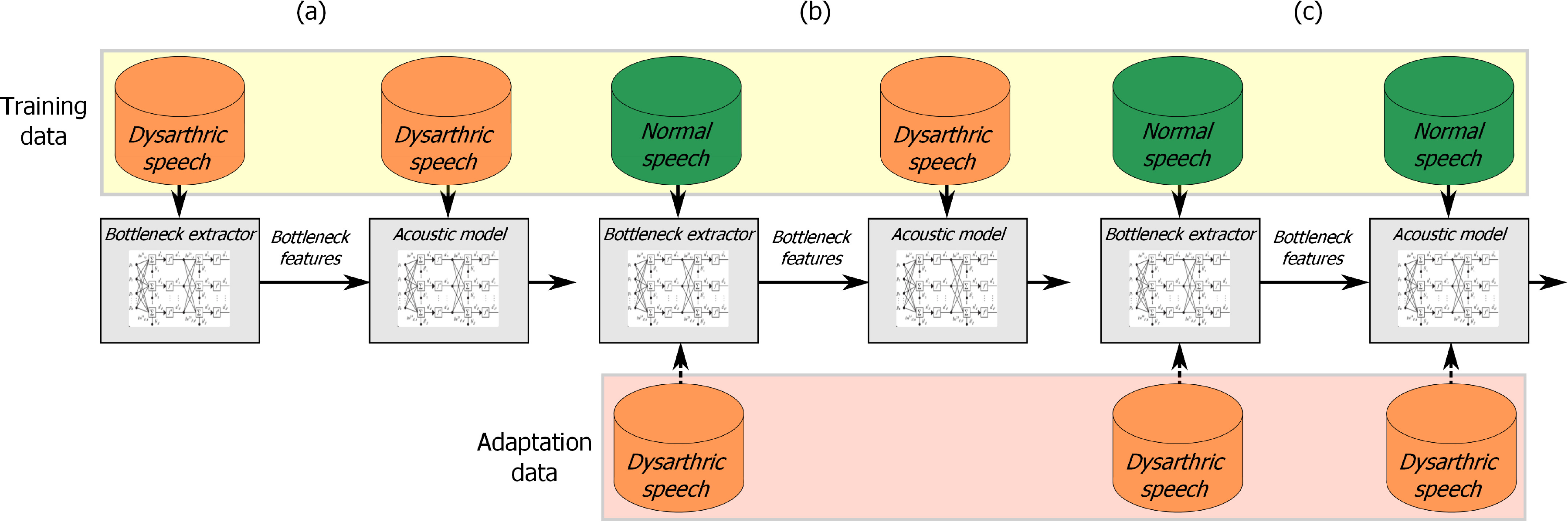}
  \caption{Different bottleneck extractor and acoustic model training strategies - (a) Training both the bottleneck extractor and acoustic model using dysarthric speech, (b) Training the bottleneck extractor using normal speech, the acoustic model using dysarthric speech and optionally adapting the bottleneck extractor using dysarthric speech and (c) Training both the bottleneck extractor and acoustic model using normal speech and optionally adapting both using dysarthric speech.}
  \label{fig:bn_str}
\end{figure*}

\subsection{Extracting articulatory features}
\label{sec:af}
The task of estimating the articulatory trajectories (in this case, the vocal tract constriction variables (TVs)) from the speech signal is commonly known as speech-to-articulatory inversion or simply speech-inversion. TVs~\cite{mitra2011,browman1992} are continuous time functions that specify the shape of the vocal tract in terms of constriction degree and location of the constrictors. During speech-inversion, the acoustic features extracted from the speech signal are used to predict the articulatory trajectories, where the inverse mapping is learned by using a parallel corpus containing acoustic and articulatory pairs. The task of speech-inversion is a well-known, ill-posed inverse transform problem, which suffers from both the non-linearity and non-unique nature of the inverse transform~\cite{richmond2001,vikramphd}.

The articulatory dataset used to train the speech-inversion systems consists of synthetic speech with simultaneous tract variable trajectories. We used the Haskins Laboratories' Task Dynamic model (TADA)~\cite{nam2004} along with HLsyn~\cite{manson2002} to generate a synthetic English isolated word speech corpus along with TVs. The English words and their pronunciation were selected from the CMU dictionary\footnote{http://www.speech.cs.cmu.edu/cgi-bin/cmudict}. Altogether 534\,322 audio samples were generated (approximately 450 h of speech), out of which 88\% of the data was used as the training set, 2\% was used as the cross-validation set, and the remaining 10\% was used as the test set. We further added fourteen different noise types (such as babble, factory noise, traffic noise, highway noise, crowd noise, etc.) to each of the synthetic acoustic waveforms with a signal-to-noise ratio (SNR) between 10 and 80 dB. Multi-condition training of the speech-inversion system has been shown to considerably improve the TV estimation accuracy under noisy scenarios, while providing comparable estimation accuracy to a system trained only using clean data \cite{mitra2017_2}. We combined this noise-added data with the clean data, and the resulting combined dataset is used for training a CNN-based speech inversion system. For further details, we refer the reader to~\cite{mitra2017}.

In this work, we use speech subband amplitude modulation features such as normalized modulation coefficients (NMCs)~\cite{mitra2014}. NMCs are noise-robust acoustic features obtained from tracking the amplitude modulations (AM) of filtered subband speech signals in the time domain. The features are Z-normalized before being used to train the CNN models. Further, the input features are contextualized by splicing multiple frames. Given the linguistic similarity between English and Dutch, we assume that the speech inversion model trained on English speech would give a reasonably accurate acoustic-to-articulatory mapping in Dutch. For a detailed comparison of the articulatory setting in Dutch and English, please see Section 21 of~\cite{collins1996}. Sivaraman et al. quantifies the performance of speech inversion under matched accent, mismatched accent, and mismatched language scenario for these two languages in \cite{sivaraman2017}.

Apart from the TVs derived from the speech inversion system trained on the synthetic dataset, we also experimented with TVs estimated from a speech inversion system trained on real speech and articulatory data. The real speech inversion system was trained on the Wisconsin X-ray microbeam dataset (XRMB)~\cite{Westbury1994a}. The pellet trajectories in the multi-speaker XRMB dataset was converted to six TVs using the method outlined in~\cite{Nam2012}. The six TVs were - lip aperture, lip protrusion, tongue body constriction location, and tongue body constriction degree, tongue tip constriction location, and tongue tip constriction degree. The XRMB data contained 46 speakers, out of which 36 were used for training and 5 each for cross validation and testing.

\subsection{Bottleneck features and model adaptation}
\label{ssec:bn-ma}

Different bottleneck extractor and acoustic model training strategies investigated in this work have been visualized in Figure \ref{fig:bn_str}. During these experiments, the DNN, CNN and TFCNN architectures described in Section \ref{ssec:acmod} are used for bottleneck extraction and acoustic modeling to compare the performance using the filterbank and articulatory features. Bottleneck extractor and acoustic model training has been performed by: (1) training a NN with a bottleneck layer, (2) extracting the bottleneck features at the output of the bottleneck layer by passing the input speech features to the trained NN and (3) training a second NN using the bottleneck features as the input. The NN obtained in the third stage is used as the acoustic model for the recognition task.

After obtaining the acoustic models and bottleneck systems, we optionally apply model adaptation as described in \cite{yilmaz2017} in explore the performance gains that can be obtained when using different bottleneck feature strategies. The idea behind such adaptation is similar to \cite{swietojanski2012, huang2013}. In these studies, considerable improvements have been reported on both low- and high-resourced languages on account of the hidden layers first trained on multiple languages and then adapted to the target language. Model adaptation is achieved by initializing the hidden layers using the NN model obtained in the first stage and retraining the model by performing extra forward-backward passes only using the available dysarthric training data with a considerably smaller learning rate compared to the one used in the first stage. The aim of this step is to tune the models on dysarthric speech as this is the type of speech to be recognized.

In previous work~\cite{yilmaz2017}, we have investigated multiple NN hyperparameters that may influence the accuracy of the final model, such as the number of retrained layers and the learning rate. Moreover, various types of speech data have been used to explore their impact on the modeling accuracy of the adapted models including accented, elderly and dysarthric speech. The best ASR performance has been obtained using only dysarthric speech in the adaptation stage. Therefore, we only use the training dysarthric speech which is only available for Dutch language in the model adaptation phase.

\section{Experimental setup}
\label{sec:expset}

\subsection{Implementation details}
\label{ssec:impdet}

We use CNNs for training speech inversion models, where contextualized (spliced) acoustic features in the form of NMCs are used as input, and the TV trajectories were used as the targets. The network parameters and the splicing window were optimized by using a held-out development set. The convolution layer of the CNN had 200 filters, where max-pooling was performed over three samples. The CNN has three fully connected hidden layers with 2048 neurons in each layer. The hidden layers have sigmoid activation functions, whereas the output layer has linear activation. The CNN is trained with stochastic gradient descent with early stopping based on the cross-validation error. The CNN hyperparameters are optimized by using a held-out development set. Pearson's product-moment correlation coefficient between the actual or ground truth and the average of all estimated articulatory trajectories are used to quantify the estimation quality. The splicing window is also optimized and a splicing size of 71 frames (355 ms of speech information) has been found to be a good choice.

For ASR experiments, a conventional context dependent GMM-HMM system with 40k Gaussians was trained on the 39-dimensional MFCC features including the deltas and delta-deltas. We also trained a GMM-HMM system on the LDA-MLLT features, followed by training models with speaker adaptive training using FMLLR features. This system was used to obtain the state alignments required for NN training. Each training set given in Table \ref{tab:data_summary} has been aligned using a GMM-HMM system trained on the same set to maximize the alignment accuracy. The input features to the acoustic models are formed by using a context window of 17 frames (8 frames on either side of the current frame). A trigram language model trained on the target transcriptions of the sentence tasks was used during recognition of the sentence tasks. This choice is motivated by the final goal of building speaker- and task-independent ASR systems, as using finite state grammars would hinder the scalability of these systems to ASR tasks with larger vocabulary.

The acoustic models were trained by using cross-entropy on the alignments from the GMM-HMM system. The 40-dimensional log-mel filterbank (FB) features with the deltas and delta-deltas are used as acoustic features which are extracted using the Kaldi~\cite{kaldi} toolkit. The NN models are implemented in Theano. The NNs trained on dysarthric Dutch training data has 4 hidden layers with 1024 nodes per hidden layer and an output layer with 5182 context-dependent (CD) states. The NNs trained on normal Dutch and Flemish data has 6 hidden layers, with 2048 nodes per hidden layer and an output layer with 5792 and 5925 CD states. These hyperparameters are chosen in the earlier stage of our previous work~\cite{yilmaz2018} based on the performance of the baseline system. The networks were trained by using an initial four iterations with a constant learning rate of 0.008, followed by learning-rate halving based on cross validation error decrease. Training stopped when no further significant reduction in cross-validation error was noted or when cross-validation error started to increase. Back-propagation was performed using stochastic gradient descent with a mini-batch of 256 training examples. The 60-dimensional bottleneck features are extracted the third layer for all neural network architectures. All ASR systems use the Kaldi decoder.

The model adaptation uses the parameters learned in the first training stage to initialize the neural net and the adaptation data is used for training with an initial learning rate of 0.001. Using a smaller initial learning rate provided consistently improved results in~\cite{yilmaz2018} for adapting different number of hidden layers. The learning rate is gradually reduced according to the regime described in the previous paragraph. The minimum and maximum number of epochs for the adaptation is 3 and 10 respectively. The same early stopping criterion is adopted during the model adaptation as the standard training.

For CNN acoustic models, the acoustic space is learned using a 200 convolutional filters of size 8 were used in the convolutional layer, and the max-pooling size was set to 3 without overlap. The TFCNN model uses 75 filters to perform time convolution, and 200 filters to perform frequency convolution. A max-pooling over three and five samples are used for frequency and time convolution, respectively. The feature maps after both the convolution operations were concatenated and then fed to a fully connected neural net. The fCNN system uses a time-frequency convolution layer to learn the acoustic space with two separate convolution filters operating on the input acoustic features. These two convolution layers are identical to the ones used in the TFCNN model. The articulatory space is learned by using a time-convolution layer that contains 75 filters, followed by max-pooling over 5 samples. 

The real AF features are extracted using a 3 hidden layer feed-forward DNN trained to map contextualized MFCC features to TVs as described in~\cite{Sivaraman2017a,Sivaraman2016}. Since the articulatory trajectories estimated by the DNN were noisy, a Kalman smoothing based lowpass filter was used to smoothen the DNN estimates. The average correlation between estimated and actual TVs on the XRMB held out test set was 0.787~\cite{Sivaraman2017a}. This speech inversion system trained on XRMB dataset was used to estimate the TVs for any utterances.

\subsection{ASR experiments}
\label{ssec:asrexp}

We investigate two different training conditions, namely with and without dysarthric training data. For the Dutch database, the ASR system is either trained on normal or dysarthric Dutch speech. Training on combination of these databases has yielded very similar results to the system trained only the normal Dutch data in the pilot experiments. Therefore, we do not consider this training setup in this paper. We further explore model adaptation in this scenario using the dysarthric training data.

For Flemish test data, we use normal Flemish and Dutch speech due to lack of dysarthric training material in this language variety. In the first setting, we only use normal Flemish speech to train acoustic models, while both normal Flemish and Dutch speech is used in the second setting motivated by the improvements reported in~\cite{yilmaz2016_2}. The recognition performance of all ASR systems is quantified using the Word Error Rate (WER).

\begin{table}[!t]
\centering
\caption{Word error rates in \% obtained on the Dutch test set using different acoustic models with mel (MFB) and gammatone (GFB) filterbank features}
\begin{tabular}{| c | c | c | c |}
\hline
AM        	&  Train.\,Data 	&  \multicolumn{2}{c|}{WER (\%)}\\
\hline
			&				&	MFB	& GFB \\
\hline \hline
DNN         &   Dys. NL     &	22.9	& 22.0	\\
\hline
CNN         &   Dys. NL    	&   21.1	& 18.8	\\
\hline
TFCNN       &   Dys. NL     &   20.3	& \bf{18.2}	\\
\hline \hline
DNN         &   Nor.\,NL     &	15.0	& 17.1	\\
\hline
CNN         &   Nor.\,NL    	&  	14.9	& 15.8	\\
\hline
TFCNN       &   Nor.\,NL     &   \bf{14.1}	& 15.9	\\
\hline
\end{tabular}
\vspace{-0.15cm}
\label{tab:res_nl}
\end{table}

\section{Results}
\label{sec:res}
In this section, we present the results of the ASR experiments performed using the acoustic models in Section \ref{ssec:acmod} trained on several speaker-independent features such as gammatone filterbanks, articulatory features and bottleneck features. Later, we apply the model adaptation approach we described in~\cite{yilmaz2017} to explore the further gains that can be obtained using a small amount of training dysarthric data. The baseline systems use DNN, CNN and TFCNN models trained on mel filterbank features and FCNN models trained using the concatenation of mel filterbank features and synthetic AFs. The details are given in~\cite{yilmaz2018}.

\subsection{Speaker-independent features}
\label{sec:sidf}
\subsubsection{Gammatone filterbank features}
\label{sec:gfbres}
The ASR results obtained on the Dutch test set using mel (MFB) and gammatone (GFB) filterbank features are presented in Table \ref{tab:res_nl}. The WERs provided by different acoustic models trained on the dysarthric Dutch speech are given in the upper panel of this table. The best ASR performance of each panel is marked in bold. Using a small amount of in-domain training data, the GFB features provide better ASR accuracy. The best-performing AM is the TFCNN with a WER of 18.2\%. The lower panel has the WERs provided by different systems trained on normal Dutch speech. In this scenario, both MFB and GFB features provide lower WERs than the previous training setting, where the best WER is 14.1\% provided by the TFCNN system trained using MFB features.

Table \ref{tab:res_vl} presents the results on the more challenging Flemish test set. In this scenario, the systems using GFB performs consistently better than the systems using MFB. Adding the Dutch data brings further improvements in the recognition of this language variety which is consistent with previous results~\cite{yilmaz2018,yilmaz2017}. Motivated by the superior performance in this scenario, the GFB features are used as the standard speech features in the rest of the experiments. 

\begin{table}[!t]
\centering
\caption{Word error rates in \% obtained on the Flemish test sets using different acoustic models with MFB and GFB features}
\addtolength{\tabcolsep}{-2.8pt}
\begin{tabular}{| c | c | c | c | c | c |}
\hline
AM        	&	Train.\,Data 	&  \multicolumn{2}{c|}{SentDys} & \multicolumn{2}{c|}{SentNor} \\
\hline
			&				&	MFB	& GFB &	MFB	& GFB \\
\hline \hline
DNN         &   Nor.\ VL    &  36.4 & 28.6 & 6.0 & 4.5 \\
\hline
CNN         &   Nor.\ VL    &  33.5 & 27.2 & 5.3 & 4.1 \\
\hline
TFCNN       &   Nor.\ VL    &  33.8 & \bf{27.0} & 5.3 & 4.2 \\
\hline \hline
DNN         &   Nor.\ VL+NL & 32.1 & 25.5 & 5.5 & 4.3\\
\hline
CNN         &   Nor.\ VL+NL & 30.1 & \bf{24.7} & 4.9 & 4.2\\
\hline
TFCNN       &   Nor.\ VL+NL & 30.1 & 25.0 & 4.9 & 4.1\\
\hline
\end{tabular}
\vspace{-0.15cm}
\label{tab:res_vl}
\end{table}
\begin{table}[!t]
\centering
\caption{Word error rates in \% obtained on the Dutch test set using different acoustic models with synthetic and real articulatory features}
\addtolength{\tabcolsep}{-2.5pt}
\begin{tabular}{| c | c | c | c |}
\hline
AM        	& Features & Train.\,Data 	&  WER (\%)\\
\hline \hline
CNN         &	GFB		&  Dys.\,NL   &  18.8 \\
\hline
TFCNN       &  	GFB		&  Dys.\,NL   &  18.2 \\
\hline
FCNN		&   GFB+synAF	&  Dys.\,NL	 &	\bf{16.7} \\
\hline
FCNN		&	GFB+realAF	&  Dys.\,NL   &	\bf{16.7} \\
\hline \hline
CNN         &	GFB		&   Nor.\,NL     &   \bf{15.8} \\
\hline
TFCNN       & 	GFB		&   Nor.\,NL     &   15.9 \\
\hline
FCNN		&   GFB+synAF	&	Nor.\,NL     &	\bf{15.8} \\
\hline
FCNN		&	GFB+realAF	&   Nor.\,NL     &	16.3 \\
\hline
\end{tabular}
\vspace{-0.15cm}
\label{tab:res_nl_af}
\end{table}
\begin{table}[!t]
\centering
\caption{Word error rates in \% obtained on the Flemish test sets using different acoustic models with synthetic and real articulatory features}
\addtolength{\tabcolsep}{-5.5pt}
\begin{tabular}{| c | c | c | c | c |}
\hline
AM        	&	Features  & Train.\,Data &  SentDys & SentNor \\
\hline \hline
CNN         &   GFB		  & Nor.\,VL    &  27.2 & 4.1  \\
\hline
TFCNN       &   GFB		  & Nor.\,VL    &  \bf{27.0} & 4.2  \\
\hline
FCNN		&   GFB+synAF	&	Nor.\,NL    &  27.5 & 4.2  \\
\hline
FCNN		&	GFB+realAF	&   Nor.\,NL    &  28.0	& 4.2  \\
\hline \hline
CNN         &   GFB		&   Nor.\,VL+NL & \bf{24.7} & 4.2 \\
\hline
TFCNN       &   GFB		&   Nor.\,VL+NL & 25.0 & 4.1 \\
\hline
FCNN		&   GFB+synAF	&	Nor.\,VL+NL & 25.0	& 4.0 \\
\hline
FCNN		&	GFB+realAF	&   Nor.\,VL+NL & 25.0 	& 4.3 \\
\hline
\end{tabular}
\vspace{-0.15cm}
\label{tab:res_vl_af}
\end{table}
\begin{table}[!t]
\centering
\caption{Word error rates in \% obtained on the Dutch test set using different acoustic models with bottleneck features}
\addtolength{\tabcolsep}{-2.5pt}
\begin{tabular}{| c | c | c | c | c |}
\hline
Train.\,BN & Train.\,AM	& DNN & CNN & TFCNN\\
\hline \hline
Dys. NL &  Dys. NL   & 15.4  & 17.1  & 16.7\\
\hline
Dys. NL &  Nor.\,NL   & 22.8  & 18.7  & 19.1\\
\hline
Nor.\,NL &  Dys. NL   & 12.9  & 13.8  & \bf{12.0}\\
\hline
Nor.\,NL &  Nor.\,NL   & 21.9  & 17.6  & 19.6\\
\hline
\end{tabular}
\vspace{-0.15cm}
\label{tab:res_nl_bn}
\end{table}

\subsubsection{Articulatory features}
\label{sec:afres}

The ASR performance of the FCNN-based ASR systems jointly using AFs and GFB are presented in Table \ref{tab:res_nl_af} and Table \ref{tab:res_vl_af}. As detailed in Section \ref{sec:af}, the AFs are extracted using speech inversion systems trained on synthetic and real speech. We refer to these features as synAF and realAF respectively. Compared to the GFB systems trained on the Dutch dysarthric training data, using AFs provides improved results with a WER of 16.7\%. Training the speech inversion system on synthetic or real speech has no significant impact on the recognition performance. A similar observation has been made on normal English speech in~\cite{Sivaraman2017a}. 

Despite the gains obtained in the systems trained on Dutch dysarthric speech, appending both synAF and realAF to GFB features of normal Dutch and Flemish training speech (cf. lower panel of Table 4 and both panels of Table 5) does not improve the recognition accuracy compared to using the GFB features only. On the Flemish test set, all ASR systems trained on the Dutch and Flemish training data provide a WER in the range of 24.7\%\,-\,25.0\% as shown in the lower panel of Table 5. The performance obtained on the control utterances spoken by the normal speakers follow a similar pattern with WERs in the range of 4.0\%\,-\,4.3\%.

\begin{table*}[!t]
\centering
\caption{Word error rates in \% obtained on the Flemish test set using different acoustic models with bottleneck features}
\vspace{-0.12cm}
\begin{tabular}{| c | c | c | c | c | c | c | c |}
\hline
Train.\,BN & Train.\,AM & \multicolumn{2}{c|}{DNN} & \multicolumn{2}{c|}{CNN} & \multicolumn{2}{c|}{TFCNN}\\
\hline
 & & SentDys & SentNor & SentDys & SentNor & SentDys & SentNor \\
\hline \hline
Nor. VL &  Nor. VL      & 29.9 & 5.4 & 30.0 & 4.8 & 30.2 & 4.8 \\
\hline
Nor. VL &  Nor.\,VL+NL   & 28.4 & 5.3 & 27.7 & 4.9 & \bf{27.5} & 4.6 \\
\hline
Nor.\,VL+NL &  Nor. VL   & 29.1 & 5.0 & 29.5 & 5.3 & 29.4 & 5.4 \\
\hline
Nor.\,VL+NL &  Nor.\,VL+NL& 28.4 & 5.2 & 28.5 & 5.4 & 28.0 & 4.8 \\
\hline
\end{tabular}
\label{tab:res_vl_bn}
\end{table*}
\begin{table*}[!t]
\centering
\caption{Word error rates in \% obtained on the Dutch test set using adapted acoustic models trained on gammatone filterbank features}
\vspace{-0.12cm}
\begin{tabular}{| c | c | c | c | c | c | c |}
\hline
Train.\,AM & Adapt.\,AM	& DNN & CNN & TFCNN & sFCNN & rFCNN\\
\hline \hline
Nor.\,NL &  -        & 17.1 & 15.8 & 15.9 & 15.8 & 16.3  \\
\hline
Nor.\,NL &  Dys.\,NL & 12.6 & 11.3 & 11.2 & 10.6 & \bf{10.3} \\
\hline
\end{tabular}
\label{tab:res_nl_gfb_ma}
\end{table*}

\begin{table*}[!t]
\centering
\caption{Word error rates in \% obtained on the Dutch test set by applying model adaptation at bottleneck extraction and acoustic modeling stages}
\vspace{-0.12cm}
\begin{tabular}{| c | c | c | c | c | c | c |}
\hline
Train.\,BN & Adapt.\,BN & Train.\,AM & Adapt.\,AM & DNN & CNN & TFCNN \\
\hline \hline
Nor.\,NL &  -        & Dys.\,NL & -        & 12.9 & 13.8 & 12.0 \\
\hline
Nor.\,NL &  Dys.\,NL & Dys.\,NL & -        & 11.8 & 13.1 & 13.4 \\
\hline \hline
Nor.\,NL &  -        & Nor.\,NL & -        & 21.9 & 17.6 & 19.6 \\
\hline
Nor.\,NL &  Dys.\,NL & Nor.\,NL & -        & 21.0 & 15.2 & 15.5 \\
\hline
Nor.\,NL &  -        & Nor.\,NL & Dys.\,NL & 11.8 & 10.4 & 10.3 \\
\hline
Nor.\,NL &  Dys.\,NL & Nor.\,NL & Dys.\,NL & 11.0 & 10.2 & \bf{10.0} \\
\hline
\end{tabular}
\vspace{-0.15cm}
\label{tab:res_nl_bn_ma}
\end{table*}

\subsubsection{Bottleneck features}
\label{sec:bnres}

The ASR results obtained using the bottleneck features on the Dutch and Flemish test sets are given in Table \ref{tab:res_nl_bn} and Table \ref{tab:res_vl_bn} respectively. We investigate different training scenarios for the bottleneck extractor and the acoustic model trained on the corresponding bottleneck features using both normal and dysarthric Dutch training data to explore which training scheme provide the best ASR performance. With a significantly large margin, the setting where TFCNN bottleneck extractor is trained on large amount of normal speech and a TFCNN acoustic model is trained on the bottleneck features of the dysarthric training speech yield the best performance with a WER of 12.0\%. The other training schemes provide inferior results compared to the results presented in Table \ref{tab:res_nl} and Table \ref{tab:res_nl_af}. 

Not having dysarthric training data for the Flemish data, we investigate using different combination with only Flemish and combined Flemish and Dutch data. The TFCNN bottleneck extractor trained on the target language and the TFCNN acoustic model trained using the bottleneck features of the combined data has a WER of 27.5\% which is better than any other training scheme. All results reported in Table \ref{tab:res_vl_bn} are still worse than the best result reported on the Flemish test set, i.e. WER of 24.7\% in Table \ref{tab:res_vl}.

\subsection{Model adaptation}
\label{sec:mares}

In the final set of experiments, we apply model adaptation to the bottleneck extractors and acoustic models using the available dysarthric training Dutch data. Table \ref{tab:res_nl_gfb_ma} demonstrates the WER provided by each adapted model trained on GFB features. Similar to~\cite{yilmaz2017}, model adaptation brings considerable improvements in the ASR performance of all acoustic models. The best result is obtained using the FCNN trained using GFB+realAFs with a WER of 10.3\%. Using GFB+synAFs yields slightly worse results compared to the GFB+realAFs with a WER of 10.6\%.

When the model adaptation applied to systems using bottleneck features, we investigate the impact of adaptation on the bottleneck extractors as well as the acoustic models. From Table \ref{tab:res_nl_bn_ma}, it can be concluded that not adapting the acoustic models trained on the normal speech has the worst performance with WERs of 19.6\% and 15.5\% without and with bottleneck extractor adaptation respectively. The last two rows of Table \ref{tab:res_nl_bn_ma} show the effectiveness of the acoustic model adaptation in this setting. Bottleneck extractor adaptation improves the performance marginally, the TFCNN system providing the best result reported on this test data with a WER of 10.0\% which is the best ASR results reported on this dataset.

\section{Summary and discussion}
\label{sec:dis}

With the ultimate goal of developing off-the-shelf ASR solutions in the clinical setting, we have described various speaker-independent ASR systems relying on designated features and acoustic models and investigated their performance on two pathological speech datasets. A Dutch and a Flemish pathological dataset with different levels of dysarthria severity, number of speakers and etiologies have been used for the investigation the described systems. The experiments on the Dutch set has given us the opportunity to investigate model adaptation, while the experiments on the Flemish test set enabled a comparison between the performance with control (normal) speech with similar spoken content. We have described multiple systems relying on novel NN architectures that are shown to more robust against spectrotemporal deviations. From the results presented in Section \ref{sec:res}, the superior performance of TFCNN compared to the baseline DNN and CNN models is clear visible in both acoustic modeling and bottleneck feature extraction scheme. Including articulatory information together with the acoustic information through fCNN models also helps against the deviations of the target dysarthric speech.  

We further compare standard acoustic features such as mel and gammatone filterbank with articulartory and bottleneck features which have been extracted using these effective NN models. Jointly using articulatory and gammatone filterbank features in conjunction with fCNN models has provided considerable improvements (cf. in Table \ref{tab:res_nl_af}) in the low-resource scenario of using only dysarthric training speech with a WER of 16.7\% on the Dutch test set compared to the 18.2\% of the TFCNN models. Using normal training speech, we do not observe improvement over the baseline CNN acoustic model on both test tests. Training the AF extractor on real or synthetic speech has not provided noticeable differences in the ASR performance on both test sets.

Bottleneck features extracted using TFCNN models give the lowest WER results (cf. in Table \ref{tab:res_nl_bn} and \ref{tab:res_vl_bn}), namely 12.0\% on the Dutch test database and 27.5\% on the Flemish test set. Moreover, using large amount of normal data for bottleneck extractor training, extracting the bottleneck features of the small amount of dysarthric speech using this extractor and finally training an acoustic model using only these features yields considerably better results compared to any other combination as shown in Table \ref{tab:res_nl_bn}.

Lastly, the model adaptation has yielded further improvements on the Dutch test set (which is the only applicable scenario due to lack of dysarthric training data in Flemish) reducing the best reported WER to 10.0\% as reported in Table \ref{tab:res_nl_bn_ma}. The system providing the best performance used the normal training speech for both the bottleneck extractor and the acoustic model training followed by the model adaptation using the dysarthric training data. Moreover, applying model adaptation to fCNN acoustic models also brings improvements with a WER of 10.3\% and 10.6\% by incorporating AFs from AF extractors trained using real and synthetic speech respectively.

The results obtained on the Flemish sentences uttered by dysarthric and control speakers enable us to quantify the performance gap between the normal and pathological speech. There is still a large gap between the ASR performance obtained on the dysarthric and control speech with a WER of 24.7\% and 4.2\% (cf. in Table \ref{tab:res_vl_af}) given by the best performing ASR on the Flemish test set. Despite the considerable WER reductions reported in this paper, the development of the generic ASR engines that are to be used in clinical procedures requires further investigation given this performance gap. 

There are multiple extensions to this work which remains as future work. One possible direction is to reduce the acoustic mismatch by including dysarthria detection/classification systems at the front-end for better understanding the nature of the speech pathology and utilizing an adapted acoustic model for recognition. In addition, data-driven (template-based) ASR approaches, which is known to preserve the complete spectrotemporal properties of speech in its templates/exemplars, can bring complimentary information when used together with the described NN architectures. Such hybrid systems relying both on statistical and data-driven approaches can enhance the recognition performance due to reduced acoustic mismatch.   

\section{Conclusion}
\label{sec:conc}
Motivated by the increasing demand for automated systems relying on robust speech-to-text systems for clinical applications, we present multiple speaker-independent ASR systems that are designed to be more robust against pathological speech. For this purpose, we explore the performance of two novel convolutional neural network architectures: (1) TFCNN which performs time and frequency convolution on the gammatone filterbank features and (2) fCNN which enables to jointly use information from acoustic and articulatory space by performing frequency and time convolution in the acoustic and articulatory space respectively. Furthermore, bottleneck features extracted using TFCNN models are compared with standard DNN and CNN models. 

Two datasets with different attributes have been used for exploring the ASR performance of the described recognition schemes. The Dutch evaluation setup with a small amount of dysarthric training speech has provided insight into effectively using normal and dysarthric training speech for training the best acoustic models on bottleneck features. Having some in-domain training data, we further apply model adaptation to all models trained on normal speech to investigate the impact of fine-tuning the neural networks on the ASR performance. On the other hand, the Flemish evaluation setup has both dysarthric and control speech of the same sentences establishing a benchmark for quantifying the performance gap between the ASR of the dysarthric and normal speech. 

From the presented ASR results, the TFCNN architecture is found to be effective both for acoustic modeling and bottleneck extraction consistently providing promising recognition accuracies compared to the baseline DNN and CNN models. Moreover, model adaptation has given significant improvements in both acoustic modeling and bottleneck extraction pipelines reducing the WERs to 10\% on the Dutch test set. Despite the performance gap reported on the Flemish test set using models trained on normal speech only, considerable WER reductions are reported on both test sets. This encourages investigation in various future directions including a dysarthria severity classification stage at the front-end of the described ASR systems and complementing these ASR systems with the information obtained from data-driven pattern matching approaches.

\bibliographystyle{elsarticle-num}
\balance
\bibliography{refs}


\end{document}